\documentclass[letterpaper, 10 pt, conference]{ieeeconf}  

\usepackage{graphicx}
\usepackage{enumerate}
\usepackage{amsmath}
\usepackage{algorithm}
\usepackage[noend]{algpseudocode}
\usepackage{subfigure}
\usepackage{chngcntr}
\usepackage{graphicx}

\RequirePackage[loading]{tracefnt}

\makeatletter
\let\NAT@parse\undefined
\makeatother

\usepackage[sort&compress,numbers]{natbib}

\usepackage{url}
\usepackage{tikz}

\usepackage{caption}

\usepackage{hyperref}
\hypersetup{
	colorlinks = true
}

\usepackage{color}

\IEEEoverridecommandlockouts

\usepackage{subfigure}
\usepackage{chngcntr}
\usepackage{multirow}
\usepackage{color}
\usepackage[sort&compress,numbers]{natbib}
\usepackage{epstopdf}
\usepackage{epsfig}
\usepackage{hyperref}

\algdef{SE}[DOWHILE]{Do}{doWhile}{\algorithmicdo}[1]{\algorithmicwhile\ #1}%

\usepackage[font=footnotesize,labelfont=bf,tableposition=top]{caption}

\usepackage{multirow}

\hypersetup{
	colorlinks = true
}

\usepackage{etoolbox}
\makeatletter
\patchcmd{\@makecaption}
{\scshape}
{}
{}
{}
\makeatletter
\patchcmd{\@makecaption}
{\\}
{.\ }
{}
{}
\makeatother

\usepackage{amsfonts}
\usepackage{amssymb}

\newcommand{\argmax}{\arg \max} 

\newcommand{\mbf}[1]{\mathbf{#1}}
\newcommand{\sbf}[1]{\boldsymbol{#1}}

\newcommand{\reals}[1]{{\mathbb R}^{#1}}

\makeatletter
\def\BState{\State\hskip-\ALG@thistlm}
\makeatother

\begin{document}
	
	\title{\Large\textbf{Intrinsically Motivated Multimodal Structure Learning}}

	\author{Jay Ming Wong${}^\dagger{}^\ddagger$ and Roderic A. Grupen${}^\ddagger$ \\
		${}^\dagger$Planning, Autonomy, and Automation Group, Draper Laboratory, Cambridge, MA, USA\\ ${}^\ddagger$Laboratory for Perceptual Robotics, University of Massachusetts Amherst, Amherst, MA, USA\\ \texttt{jmwong@draper.com, grupen@cs.umass.edu} 	
	}

	\maketitle
	
	\begin{abstract}
		
		We present a long-term intrinsically motivated structure learning method for modeling transition dynamics during controlled interactions between a robot and semi-permanent structures in the world. In particular, we discuss how partially-observable state is represented using distributions over a Markovian state and build models of objects that predict how state distributions change in response to interactions with such objects. These structures serve as the basis for a number of possible future tasks defined as Markov Decision Processes (MDPs). The approach is an example of a structure learning technique applied to a multimodal affordance representation that yields a population of forward models for use in planning. We evaluate the approach using experiments on a bimanual mobile manipulator (uBot-6) that show the performance of model acquisition as the number of transition actions increases. 
		
	\end{abstract}
	
	\section{Introduction}
	\label{sect:intro}
	
	
	Humans accumulate a large repertoire of action-related knowledge from experiences over a lifetime of problem solving. As infants, we explore the world and entities in our environment, building representations for future use through play. We do this because we are inherently curious and discovery is rewarding for its own sake---we are \emph{intrinsically motivated} to acquire models of the world \cite{Chentanez2005,Itti2006,Singh2010}. Many researchers have explored intrinsic motivation as a key component for developing curious, exploratory, and autonomous behavior---for instance, in the acquisition of visuomotor skills for robots \cite{Oudeyer2007}. Hierarchical approaches have been developed employing intrinsic motivation to learn new skills autonomously \cite{Utgoff2002, Barto2004}. A number of intrinsic motivators have been proposed \cite{Oudeyer2007,Hart2009} with approaches in contrast with previous work that relied on hand-built representations tailored to a particular task \cite{Nilsson1984,Allen1985,Grimson1987}. Our view is that autonomous exploration and intrinsically motivated discovery should prove more robust and transferable than hand built knowledge representations. 
	
	Insight from cognitive psychology has influenced many researchers to investigate knowledge representations  \cite{Natale2004,Stoytchev2005,Lopes2007,Montesano2008,Montesano2009,Hart2009,Detry2010,Kruger2011,Ku2014,Ku2014b,Sen2014,Ku2015b,Ugur2014,Song2015,Worgotter2015,Hester2015,Ugur2015,Kaiser2015a}. Among these, the notion of \emph{direct perception} and \emph{affordances} proposed originally by Gibson \cite{Gibson1977} is particularly relevant to our approach. Gibson's theory of affordance advocates for modeling the environment directly in terms of the actions it affords. These representations are idiosyncratic and reflect only those actions that can be generated by the agent. Research has been done to investigate the autonomous acquisition of such affordance representations with intrinsic motivators. For instance, an example of multiple intrinsic reward functions have been proposed to learn the transition dynamics of a particular task \cite{Hester2015}. Others have looked into domain-independent intrinsic rewards, like novelty or certainty, for learning adaptive, non-stationary policies based on data gathered from experience \cite{Hart2009,Sequeira2014}. In particular, \emph{model exploration programs} have been presented \cite{Hart2009}, but methods reported to date lack multimodal sensor integration and do not produce knowledge structures that are easily transferrable to other tasks. 
	
	A number of studies have presented methods to learn affordance representations through imitation \cite{Lopes2007,Montesano2008}, building experience-grounded representations called \emph{Object--Action Complexes} (OACs) \cite{Kruger2011}. Affordance models such as OACs provides a basis for \emph{structural bootstrapping}, allowing existing knowledge to generalize to otherwise unexplored and novel tasks and domains \cite{Worgotter2015}. Such generalizability may be used to support planners that use learned representations as \emph{forward models} $f: a,s\mapsto s'$ which is synonymous to state transition models in MDPs. However these models do not necessarily encode the prerogatives of the embodied system nor can they be easily adapted to other robots \cite{Detry2010}. Moreover, they require the guidance of a teacher and are relatively cumbersome, though methods have been proposed with intrinsic motivators but assume predefined structure prior to training \cite{Ugur2014}. 
	
	This paper adopts a different form of affordance representation that is lighter weight, and thus, better serves planners that need to roll out a number of these forward models during planning. In fact, this representation encodes only essential Markovian components concerning information regarding states, actions, and transition dynamics, $s,a \mapsto s'$, and thus, can be reused generally for all tasks that can be formulated as MDPs. The main contribution of this paper is the presentation of an intrinsically motivated structure learning approach that builds complete action-related representations of objects using multimodal percepts. The resulted is called an \emph{Aspect Transition Graph} (ATG) model. Previous planning architectures using hand built versions of these models have been successful, however, this paper contributes a structure learning approach to acquiring them autonomously. 
	
	We present the first autonomously learned ATG representation with continuously parametrized action edges in the literature. These representations can be used to serve as forward models in belief-space planning infrastructure on real robot systems \cite{Grupen2016}. A number of studies have integrated ATG affordance representations into the model base as a fundamental attribute in the model-referenced belief-space planning architecture. For instance, Sen showed that the \emph{object identification task}\footnote{\emph{Object identification task}---a robot is given a large corpus of models it has interacted with in the past in memory and a real-world object and is asked to identify which model the object belongs to. Work of this nature falls into the \emph{active vision} field which suggests that vision along can not solve such tasks, but the embodied system must execute actions to condense its belief towards the correct object model in memory.} can scale up to 10,000 object models by pruning those with insufficient support \cite{Sen2014}. Though models used in studies like those of Sen \cite{Ku2014,Ku2014b,Sen2014,Ku2015b} do not inherently encode transition dynamics learned by the robot and therefore are not robust to unexpected outcomes without inherently encoding the system's uncertainty into the representations. Work by Ku \emph{et al.} has shown that the ATG structure is, however, capable of fine grain error detection, in which surprising outcomes that are not encoded in the object model trigger a \emph{What's up?} action \cite{Ku2015b}. Transition properties in our work are all learned autonomously, encoding unknown properties of the underlying system, and thus are robust to errors. 
	
	
	%
	%
	%
	
	\section{Technical Approach}
	This manuscript presents an algorithm for autonomous structure learning incorporating multiple sensor modalities and robot actions to produce lasting artifacts that can be used in the future (i.e. for reasoning and planning). The algorithm is presented in the context of a system that builds object representations for future use, but may span a number of other domains that require learning inherent structure in a task independent manner.
	
	\subsection{Affordance Representation}
	
	\begin{figure}
		\setlength{\unitlength}{1pt}
		\begin{picture}(250,95)(0,0)
		
		\put(12,0) {\centering \includegraphics[width=0.45\textwidth]{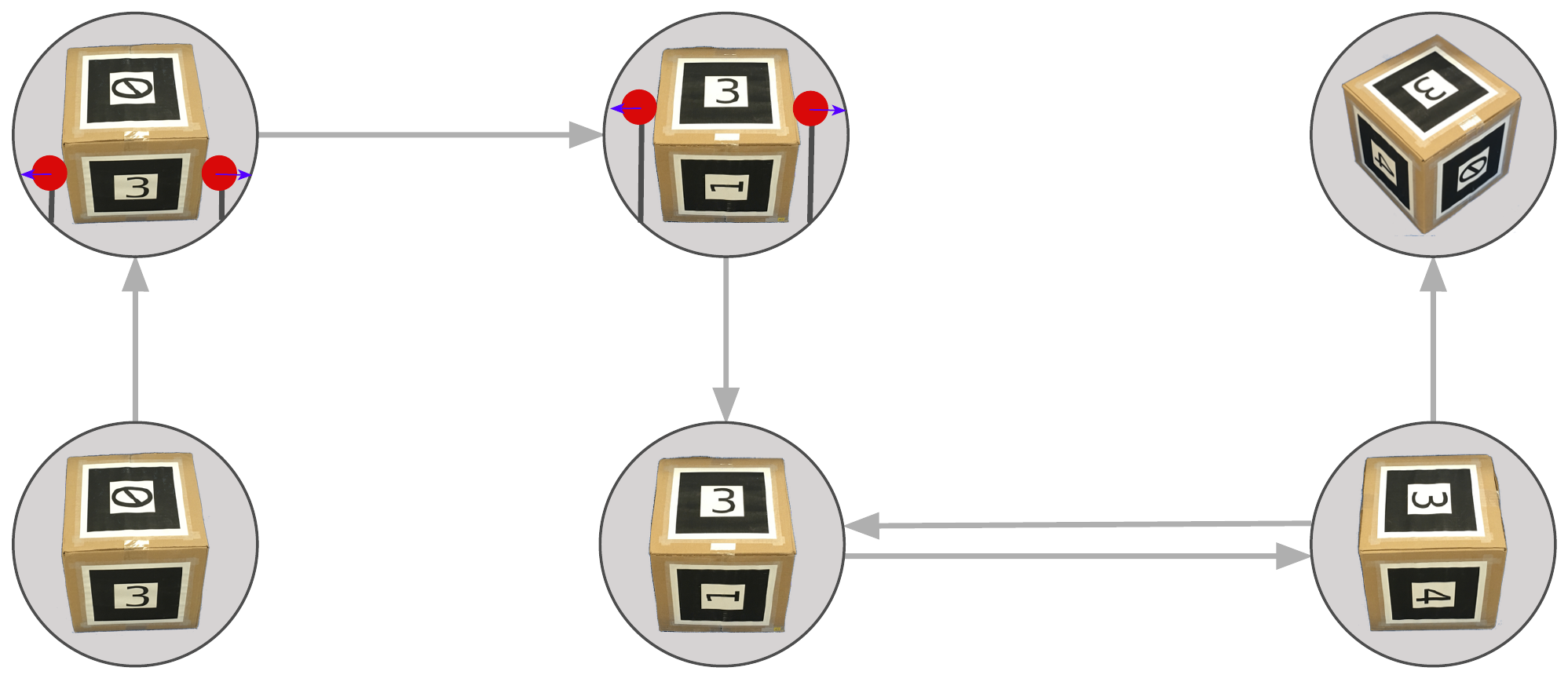}}
		\put(0,47) {\textsc{Grasp}($\mu_{xyz}, \Sigma$)}
		\put(64,83) {\textsc{Lift}}
		\put(195,47) {\textsc{Orbit}($\frac{\pi}{4}, \sigma$)}
		
		\put(145,10) {\textsc{Orbit}($\pi,\sigma$)}
		\put(145,25) {\textsc{Orbit}($-\pi, \sigma$)}
		\put(105,50) {\textsc{Place}}
		\end{picture}
		\caption{An example of a \emph{partially} constructed Aspect Transition Graph where affordances of a die-like object are encoded in \emph{aspect nodes} connected by directed edges representing actions. The labels on the edges here correspond to possible \emph{control programs} with parameters that result in successful transitions. The blue arrows at the hand for two of these nodes indicates tactile information. 
		}
		\label{fig:partial-atg-example}
	\end{figure}
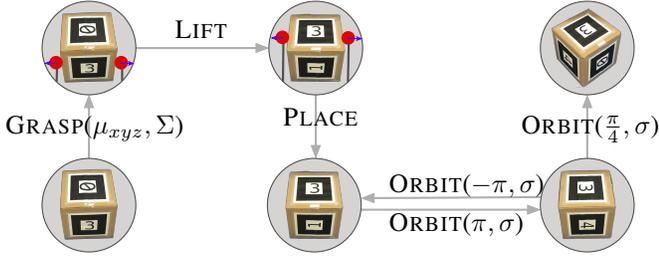
	
	Our approach encodes affordances in a graphical structure called an \emph{Aspect Transition Graph} \cite{Ku2014b}---which is defined as a directed multi-graph $G=(\mathcal{S}, \mathcal{A})$ where $\mathcal{S}$ denotes a set of \emph{aspect nodes} connected by action edges $\mathcal{A}$. A pictorial example of a partial ATG is illustrated in Figure~\ref{fig:partial-atg-example} describing several plausible interaction outcomes with a particular object. Sensory information in multiple modalities (vision and touch) is integrated into the aspect nodes in the graph. Each parameterized action $a\in \mathcal{A}$ uses a learned search distribution for motor references that reliably transition between aspects. References are defined to be (multivariate) Gaussian distributions $\mathcal{N}(\mu,\Sigma)$ in Cartesian space describing the areas in object frame where the robot has successfully detected a target perceptual reference from this initial state in the past. An \emph{aspect node} is a state representation defined as a geometric constellation of features derived from multiple sensor modalities. For example, an aspect node may be a geometric constellation of visual features present in a particular ``field of view.'' In theory, the number of features for any given object may be arbitrarily populous. As a result, the size of an ATG representation for the object, may be very large. However in principle, aspects encode \emph{affordances} and are bounded by the number of actions $|\mathcal{A}|$ an embodied system may perform on the object that changes the relative sensor geometry. Aspect nodes are stored in the model as pointers to specific features that are indexed and arranged by type and value ordered chronologically by discovery time. Each feature is defined by three fields (id, type, value) in addition to a mean and covariance describing the likely Cartesian positions $\mu \in \reals{3}, \Sigma \in \reals{3\times3}$ in object frame\footnote{For instance, the bottom leftmost aspect node in Figure~\ref{fig:partial-atg-example} could be defined as a 0-1 aspect, pointing to feature id: 0 of type: `ARtag' and value: `3' and feature id: 1 of type `ARtag' and value: `0'.}. 
	
	%

	\subsection{Aspect Observation}
	\label{sect:aspect-observation}
	Performing an observation of the scene creates a feature list. Observations consist of maximum likelihood Cartesian features derived from a Kalman filter that summarizes the history of observations to this point in terms of a mean observation and an associated spatial covariance. The current aspect $s$ is obtained by observing the multimodal features in the scene and applying some mapping $\mathcal{F}: f_i, f_j, \hdots, f_n \mapsto s$ which defines the aspect as a subset or encoding of relevant features in the feature list. This paper implements $\mathcal{F}$ by simply returning the string representation of the geometric (order-specific) collection of all features over all modalities present, obeying some aspect geometry. Future work hopes to extend $\mathcal{F}$ to incorporate a generalized Hough transform to vote for the position of the model coordinate frame \cite{Grupen2016}. 
	
	
	
	\subsection{Control Actions}
	Each edge in the ATG is a closed-loop controller $\phi|^\sigma_{\tau}$ that combines potential functions ($\phi \in \Phi$) with sensory ($\sigma \subseteq \Sigma$) and motor resources ($\tau \subseteq \mathcal{T}$) using the \emph{Control Basis} framework \cite{Huber1996,Sen2014}. Such controllers achieve their objective by following gradients in the potential function $\phi(\sigma)$ with respect to changes in the value of the motor variables $\sbf{u}_\tau$, described by the error Jacobian $\mbf{J}=\delta\phi(\sigma)/\delta \sbf{u}_\tau$. References to low-level motor units are computed as $\Delta \sbf{u}_\tau = \kappa \mbf{J}^\# \Delta\phi$, where $\kappa > 0$ is a small gain, $\mbf{J}^\#$ is the pseudoinverse of $\mbf{J}$, and $\Delta \phi$ is the difference between the reference and actual potential. For the method proposed in this paper, it is assumed that the number of actions $|\mathcal{A}|$ and their parameter spaces are known \emph{a priori}. 
	
	\subsection{Action Selection}
	\label{sect:action-selection}
	Affordances of a given object can be determined by exploring actions that cause a transition in the aspect space. The approach presented in this manuscript selects actions to learn affordance representations---actions are selected for execution to achieve two kinds of reward:
	\begin{enumerate}
		\item \textbf{Discovering novelty} finding new aspect nodes $s\in \mathcal{S}$ and revealing new aspect transitions
		\item \textbf{Refining parameters} $\rho\sim\mathcal{N}(\mu,\Sigma)$ for actions $a \in \mathcal{A}$ encoded in search distributions of known transitions $p(s'|s, a(\rho))$ between aspect nodes $s$ and $s'$ 
	\end{enumerate}			
	To encourage coverage over the space of actions, a Latin Hypercube Sampled (LHS) space (over the domain of each action $a\in\mathcal{A}$) is introduced for each aspect node $s$. Action parameters $\rho$ are randomly sampled during exploration---first by type, then by parameters defined in the LHS-space. \emph{Parameter refinement} at some aspect node $s$ is addressed by querying and evaluating all outgoing edges $\forall a \in \mathcal{A}: p(s'|s, a) > 0$ and sampling parameters $\rho$ for the \emph{highest-valued} action $a^*$.  It is important to note that intrinsic reward alone is insufficient for producing a complete ATG representation since it selects actions that exploits learned structure, however, structure must first be discovered hereby stressing coverage. 
	
	\subsection{Affordance Modeling and Intrinsic Reward}
	%
	%
	%
	%
	%
	
	\label{sect:intrinsic-motivation}
	The \emph{highest-valued} action $a^*$ is a property of intrinsic motivation that drives the affordance modeling construction process. This is performed by storing and updating aspect node structure and transition information given learning experiences in the form of $\langle s, a, \rho, s'\rangle$, which is achieved by obtaining the aspect definitions $s$ and $s'$ (outlined in Section~\ref{sect:aspect-observation}) respectively before and after the selection and the execution of the action $a$ with parameters $\rho$ (from Section~\ref{sect:action-selection}). In short, the robot observes the current state, performs actions, and memorizes the new state produced. With each experience example, the parameter $\rho$ is added to a nonparametric distribution along the action edge $a$ in the ATG corresponding to the transition from $s$ to $s'$. This nonparametric data structure stores all the robot's past experiences and describes all control parameters $\rho$ that result in a particular perceptual outcome. Next, a new Gaussian distribution is generated using the set of all $\rho$ existing in $a$ and intrinsic reward $r(s,a(\rho), s')$ is computed.
	
	Q-value iteration \cite{Watkins1992,Sutton1998} is implemented to establish a value function where at any time, the highest value corresponds to regions of interest in the parameter-space with high uncertainty. The intrinsic reward function uses the difference in the variance of experiences achieved from the same state under the same action as originally proposed in \cite{Hart2009}. For use with the ATG representation, the intrinsic reward takes a slightly different form,
	\begin{align*}
	r(s,a(\rho), s') = \mbox{abs}(||\Sigma_{k}||_2 - ||(\Sigma_{k-1})||_2)
	\end{align*}
	where $k$ indicates the value after the $k$th experiment, $\Sigma_k$ refers to the sample variance in the Gaussian $\mathcal{N}(\mu,\Sigma)$ that describes the action parameter distribution that consists of all parameters $\rho$ that accomplishes transition $s, a(\rho)\mapsto s'$. Then, $\Sigma_{k-1}$ is the Gaussian distribution that does not include the most recent action $a(\rho_k)$. In the case that the edge $a$ (corresponding to $s\rightarrow s'$) is novel or the aspect $s'$ is novel, a new transition edge or node is created---reward is then the differential variance between an arbitrarily wide Gaussian and an arbitrarily thin Gaussian centered at $\rho$. In practice, the reward is bounded by some defined maximum, for instance $r(s, a(\rho),s')=1.0$. The key insight for using an intrinsic reward function of this form is that it encourages the \emph{consumption} of reward through actions. In other words, it promotes the selection of actions that produce a high differential variance. As these distributions converge, intrinsic reward diminishes, hence encouraging other action parameters contributing to other transitions to be selected. 
	
	In addition to this update in the transition properties, all features that define the new aspect $s'$ are either added or updated as appropriate. In summary the overall algorithm is described simply as,
	\begin{algorithm}
		\caption{Multimodal Structure Learning}\label{alg:msl}
		\begin{algorithmic}[1]
			\State $f \gets $ \mbox {\textsc{Nil}}
			\Do
			\State $a, \rho \gets$\mbox{Select params by LHS or $\argmax_{a} f(s,a, s')$} 
			\State \mbox{Do $a, \rho$ and obtain experience $\langle s, a, \rho, s'\rangle$}
			\State $r(s,a(\rho), s') \gets \mbox{abs}(||\Sigma_{k}||_2 - ||(\Sigma_{k-1})||_2)$
			\State \mbox{Update value $f(s,a,s')$ with reward $r(s,a(\rho),s')$}
			\State \mbox{Update ${}_{}\mathcal{N}_{s,a \rightarrow s'}$ with current action params $a(\rho)$}
			\doWhile{$f(s,a,s') > \varepsilon: \forall s,a,s' $}
		\end{algorithmic}
	\end{algorithm}

	\section{Experiment Methodology}
	
	\subsection{Robot Platform}
	
\begin{figure}
		\centering
		\includegraphics[width=0.46\textwidth]{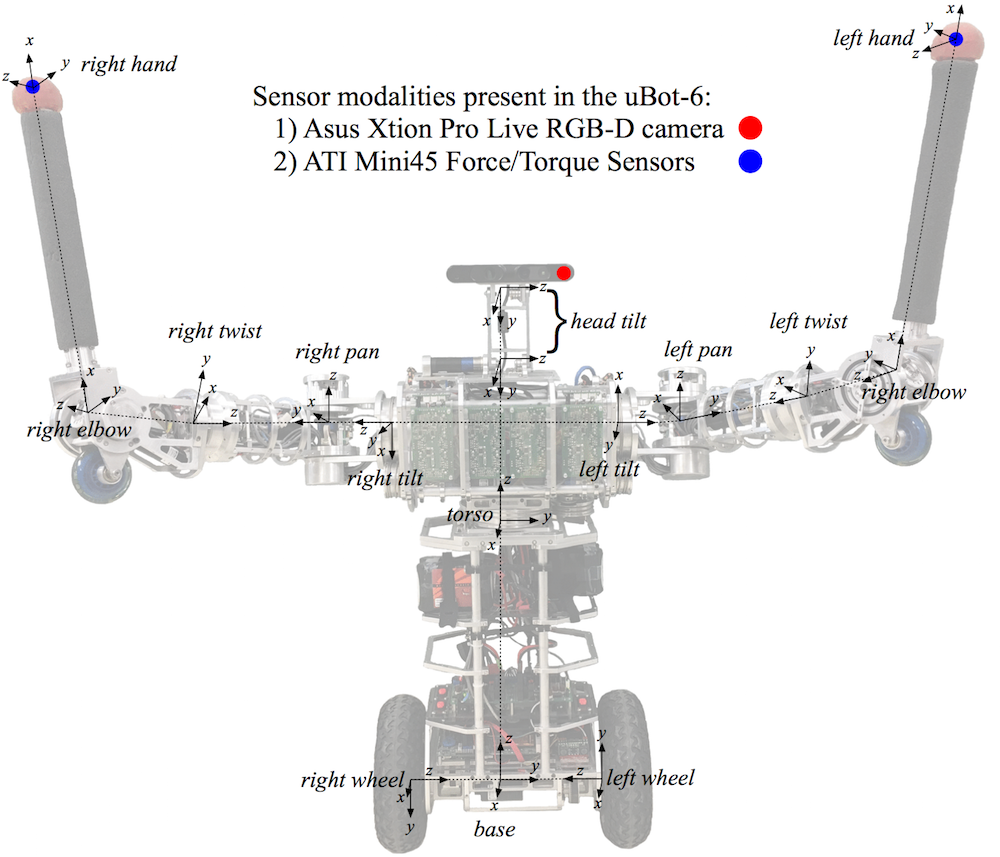}
		\caption{The uBot-6 mobile manipulator has multiple degrees of freedom (DOF) supporting the ability to solve a number of tasks in many different ways. Control programs engage subsets of these DOF illustrated in the kinematic chain when executing actions in an attempt to search for intrinsic reward. Visual and tactile sensors allow for the robot to perform perceptual actions. (Best viewed in color)}
		\label{fig:uBot-6}
	\end{figure}

	
	Experiments are done on a dynamic simulation of the uBot-6 platform, a 13 DOF, toddler-sized, dynamically balancing, mobile manipulator \cite{Ruiken2013}  equipped with an Asus Xtion Pro Live RGB-D camera (shown in Figure~\ref{fig:uBot-6}) and two ATI Mini45 Force/Torque sensors one in each hand (not shown in figure). Control actions are executed by the robot to establish new sensor geometries and reveal new aspects. Collectively, experimental results compile a total of over $250$ hours of robot simulation.
	
	\subsection{Sensor Modalities (Features and Aspects)}
	The Asus RGB-D camera and ATI Mini45 Force/Torque sensors provides visual and tactile information to the robot. Visual features are extracted and ordered such that the feature list is populated with priority on feature Cartesian location (left to right, bottom to top of the image). Primitive tactile features consist of the contact force $\hat{\sbf{f}} \in \reals{3}$, from which the sum of squared contact forces $\sum_{i=L,R} \sbf{f}_i^T \sbf{f}_i$ and sum of squared contact moments $\sum_{i=L,R} (\sbf{r}_i \times \sbf{f}_i)^T((\sbf{r}_i \times \sbf{f}_i)$ are computed at the centroid of the pair of contacts
	measured, where $L$ and $R$ signify left and right, respectively. Bimanual grasp configurations where the squared force and moment residuals are minimized
	simultaneously are considered to be valid grasp hypotheses. This form of tactile information is added to the visual components of an aspect to obtain a multimodal {\it aspect node} which we propose as a representation of the {\it interaction state}. 
	
	\subsection{Control Programs (Actions)}
	The set of actions $\mathcal{A}$ in these experiments consists of two control programs: \textsc{Orbit} and \textsc{Grasp}. These actions are responsible for changing the relative sensor geometries of the robot relative to the object and, as a result, cause probabilistic transitions to new aspect nodes. 
	
	\textsc{Orbit} is a \emph{locomotive control program} that changes the viewpoint geometry of the Asus Xtion Pro Live RGB-D camera. As shown in Figure~\ref{fig:uBot-6}, a number of motor resources (or degrees of freedom) may be used to achieve this. Our approach implements the translational and rotational axis of the base subject to the constraint that the final heading is toward the object. The action is parametrized by an angle $\theta$ about the world $\hat{\sbf z}$ axis with some fixed orbit radius $r$ defined a priori. In the experiment presented in this paper, $r=1.0$ ($m$).
	
	
	\textsc{Grasp} is a control program that changes the sensor geometry of the ATI Mini45 Force/Torque sensors in each hand to some new location in the scene. The grasp action engages the mobility resources, if necessary, to put the object within reach and then engages the arms to place the hands at Cartesian goals relative to the object where compressive forces are applied. This experiment is used to update the search distributions for mobility goals and hand placement based on the percept $\sum \sbf{f}_i = \sum \sbf{m}_i = 0$. A model is acquired that can be used to transform a partial observation of an aspect node, perhaps composed exclusively of visual features, into new aspect nodes that assert that the grasp force and moment objectives are also met (i.e. that an adequate grasp configuration exists). Each \textsc{Grasp} action is paired with a \textsc{Release} counterpart in which the robot releases the held object and retreats to the previous pre-grasp base Cartesian position. 
	

	\subsection{Target Objects}
	Since the approach presented in this paper makes no assumption regarding the underlying object and only concerns the aspects that are afforded, it can theoretically be applied to any object. However, in our validation experiments we use a simple object geometry as a proof of concept whose ATG can be evaluated. In all experiments, the uBot-6 is presented with an ARcube object in Gazebo with a random configuration. ARcubes are rigid $29$ cm cubes with a single ARtag mounted on each of the six faces\footnote{An open-source ARToolKit is available (\texttt{http://artoolkit.org}) for detecting and localizing the tags.}. Visual observations of these features establish the location of the center of each tagged face. 
	
	The ARcubes in this experiment provides a form of validation since the number of aspects and transitions of ARcubes are enumerable by hand. Further, the transition parameters for \textsc{Orbit} on this specific object tends to be at about an interval of $\pi/4$---this intuition can be used to verify the correctness of models produced by our approach by establishing a ground truth representation. 
	
	
	\subsection{Experiment Layout}
	
	The first experiment is a necessary validation step in which the proposed approach presented in this paper is compared against a base-line approach. Methods like \cite{Lopes2007,Montesano2008,Ku2014b} adopt either imitation or memorization paradigms for the construction of affordance models, with \cite{Ku2014b} being the only work specifically for ATGs. Work to learn ATGs have been mainly accomplished by learning through demonstrations, thus are not truly autonomous. Here, we present a base-line method in which the robot randomly explores control parameters, observes the scene, and memorizes its effects in terms of aspect transitions. Such a method is guaranteed to converge to a complete affordance model given sufficient time and serves as a valid contender for comparisons. Both the proposed and base-line methods are compared against a ground truth model with only the \textsc{Orbit} action for validation. 
	
	The second experiment aims to inspect the result in which additional sensor modalities and actions are introduced. In the first experiment, the action space $\mathcal{A}$ consisted solely of parametrized \textsc{Orbit} actions and only the Asus camera existed in the set of sensor modalities. The uBot-6 has access to both \textsc{Orbit} and \textsc{Grasp} actions and visual and tactile features from the Asus RGB-D camera and ATI force/torque sensors. 
	
	\section{Results}
	
	The results presented in the first experiment contains over $150$ hours of simulation, consisting of five trials for each approach. The affordance model corresponding to ground truth has eight visual aspect nodes and $64$ interconnected transition edges in the ATG. Error (in radians) is computed by the absolute difference between the learned model and the ground truth for the means of the distribution along all transition edges. The error related to each edge is then averaged for an overall model error. If an edge is not discovered by the learning method the error for that edge is set to the maximum, $\pi$. Table~\ref{tab:orbit-compare} lists the error for both the proposed and the random memorization approaches after a specific number of actions. In all cases, the proposed method achieves lower errors and in many of these cases, the difference is statistically significant ($p<0.05$). It is also evident that the proposed approach is capable of acquiring more accurate affordance representations faster and more reliably (with significantly lower standard deviation). 
	\begin{table}
		\centering
		\begin{tabular}{|c|c|c|c|}
			\hline
			$|\mathcal{A}|$           &  \textsc{p-value} & \textsc{Proposed} & \textsc{Rand+Memorize}\\ \hline
			$50$       &   	$0.0034$		&   	$1.4934\pm0.0732$	 &  $1.7551\pm0.2097$    \\ \hline 
			$100$       &   	$0.0063$		&   	$0.7598\pm0.1477$	 &  $1.0919\pm0.2929$    \\ \hline 
			$150$       &   	$0.0068$		&   	$0.4242\pm0.1062$	 &  $0.7666\pm0.2876$    \\ \hline
			$200$       &   	$0.0125$		&   	$0.2701\pm0.0630$	 &  $0.5797\pm0.3345$    \\ \hline 
			$250$       &   	$0.0071$		&   	$0.2180\pm0.0648$	 &  $0.5474\pm0.3435$    \\ \hline 
			$300$       &   	$0.0153$		&   	$0.1574\pm0.0409$	 &  $0.4626\pm0.3511$    \\ \hline 
			$350$       &   	$0.0217$		&   	$0.1447\pm0.0258$	 &  $0.4388\pm0.3511$    \\ \hline
			$400$       &   	$0.0254$		&   	$0.1251\pm0.0187$	 &  $0.4197\pm0.3608$    \\ \hline  
			$450$       &   	$0.0251$		&   	$0.1219\pm0.0177$	 &  $0.4071\pm0.3511$    \\ \hline 
			$500$       &   	$0.0323$		&   	$0.1112\pm0.0060$	 &  $0.3865\pm0.3469$    \\ \hline 
		\end{tabular}			
		\vspace{5pt}
		\caption{Model error comparison between the proposed structure learning approach and a base-line approach where transition probability from $s \stackrel{a}{\rightarrow} s'$ is approximated by randomly exploring action $a$ from every initial state $s$ and recording an estimated $p(s' | s, a)$ for all $s’$ with only the \textsc{Orbit} control program in the action set. Evaluations are performed against an empirical model taken as ground truth and errors correspond to the average error (in radians) over all transitions in the model.}
		\label{tab:orbit-compare}	
	\end{table}
	
	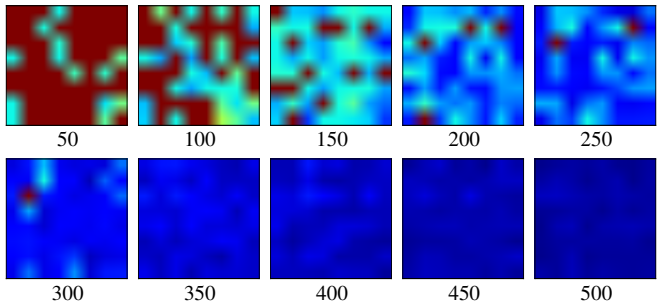
\begin{figure}
		\vspace{-20pt}
		\centering
		\begin{picture}(250,112)(0,0)
		\put(0,60) {\centering \includegraphics[width=47px, height=47px]{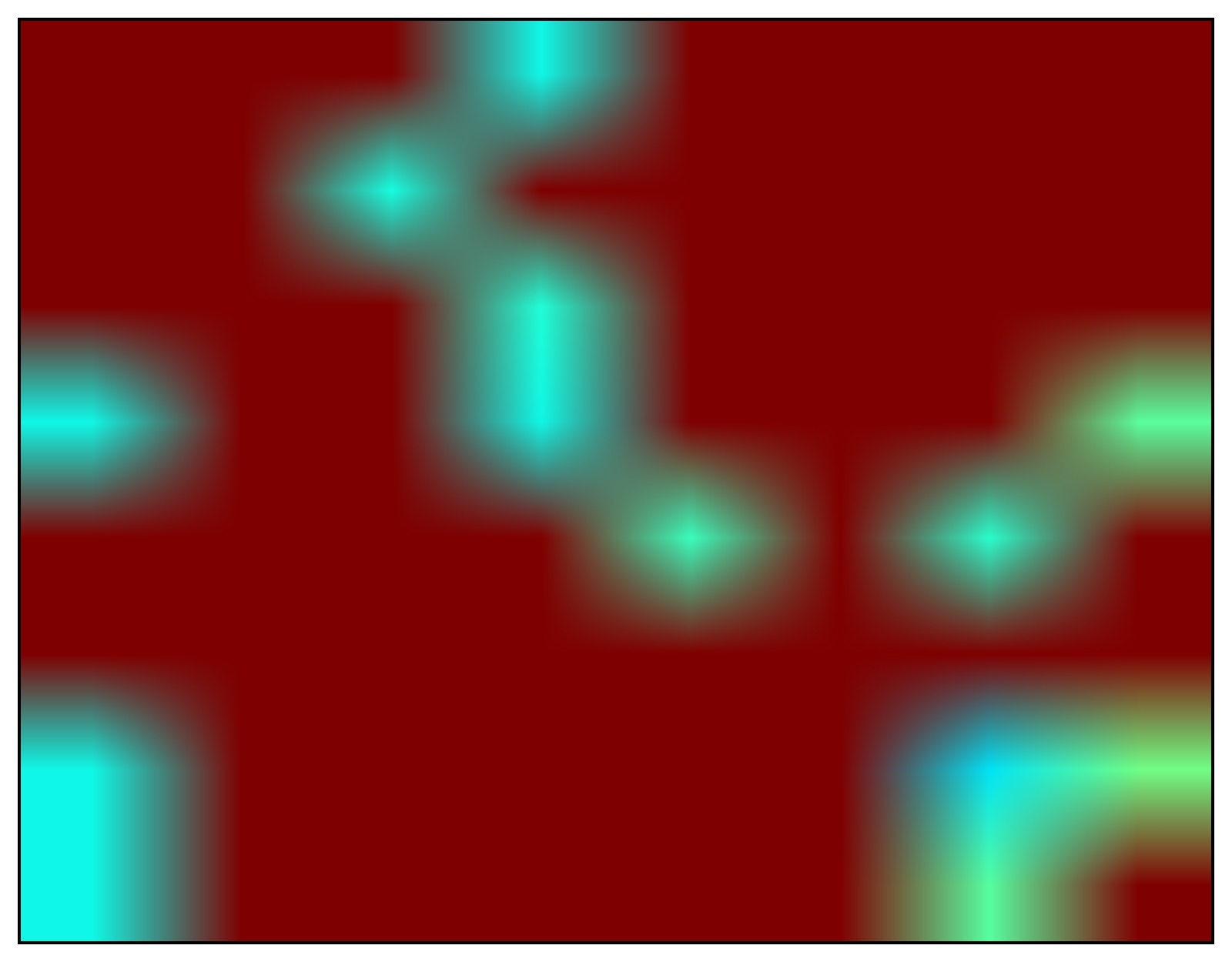}}
		\put(50,60){\centering \includegraphics[width=47px, height=47px]{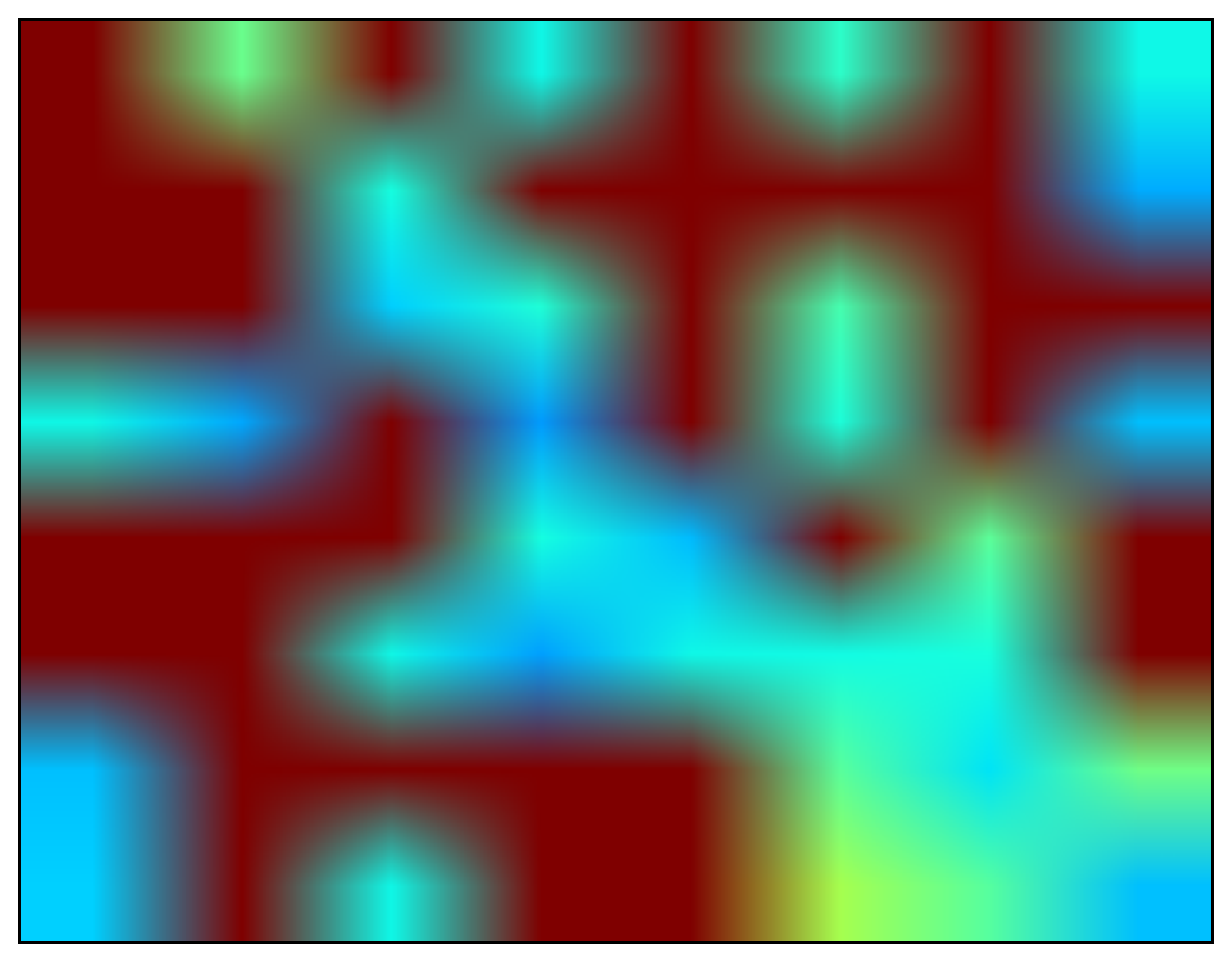}}
		\put(100,60) {\centering \includegraphics[width=47px, height=47px]{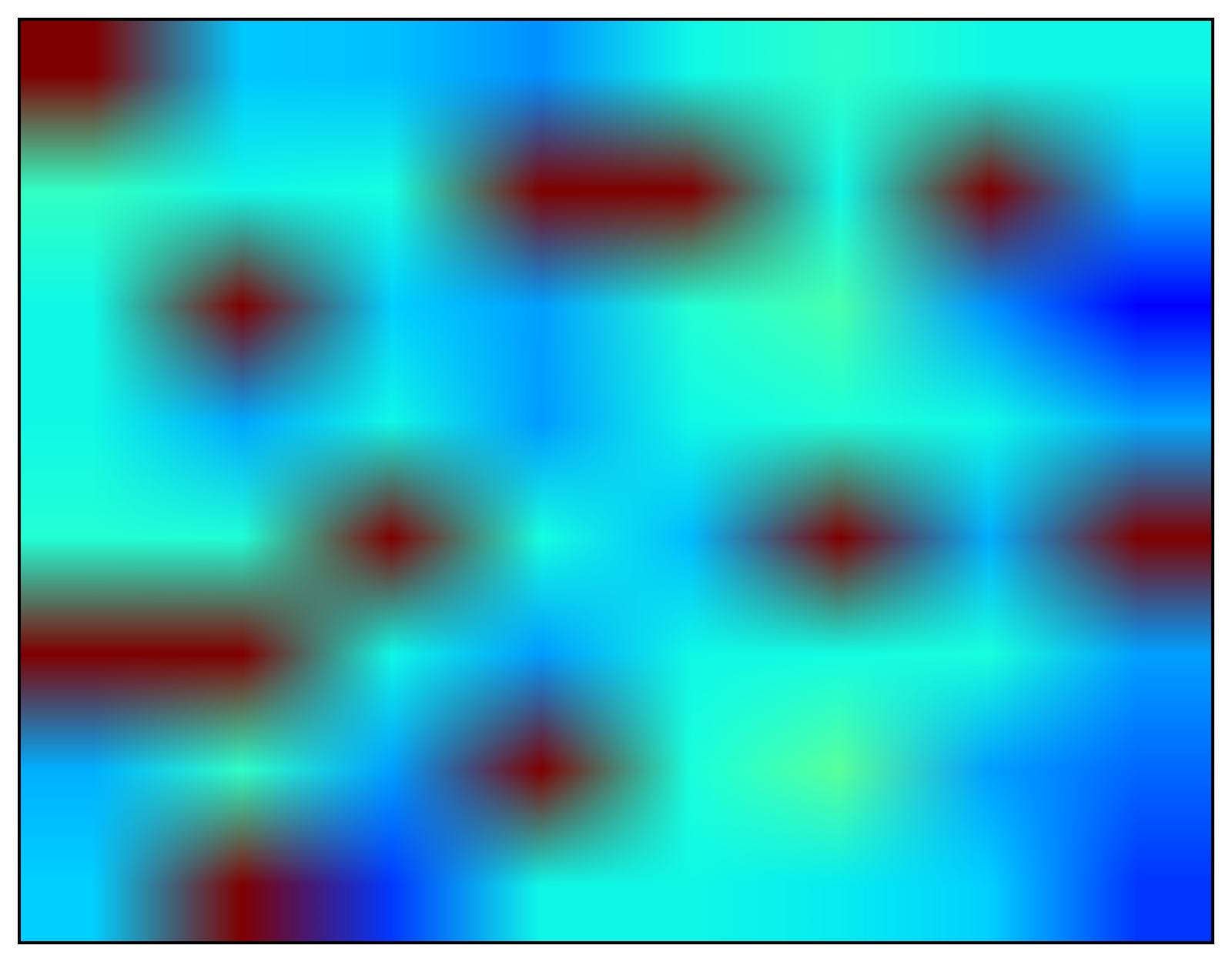}}
		\put(150,60) {\centering \includegraphics[width=47px, height=47px]{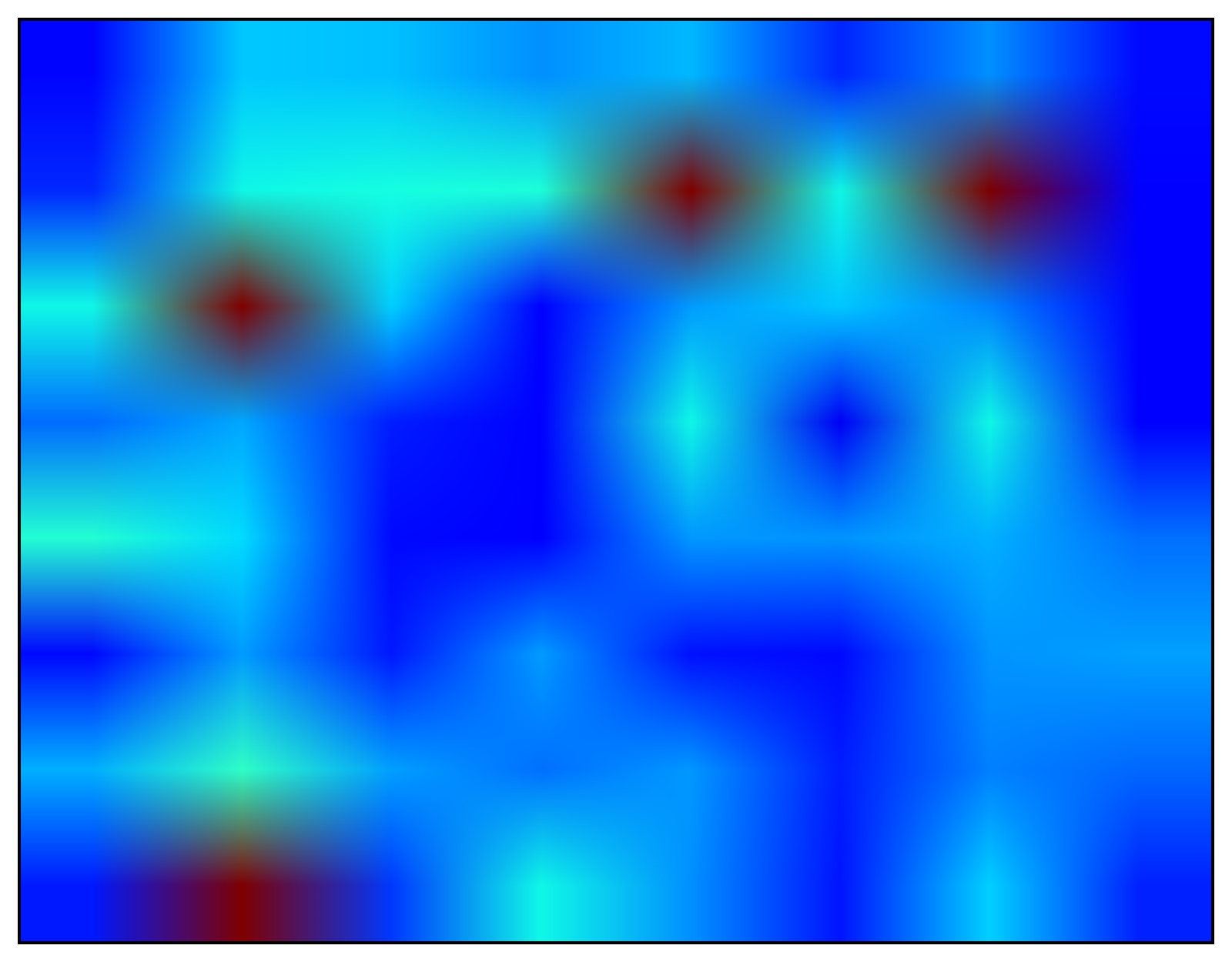}}
		\put(200,60) {\centering \includegraphics[width=47px, height=47px]{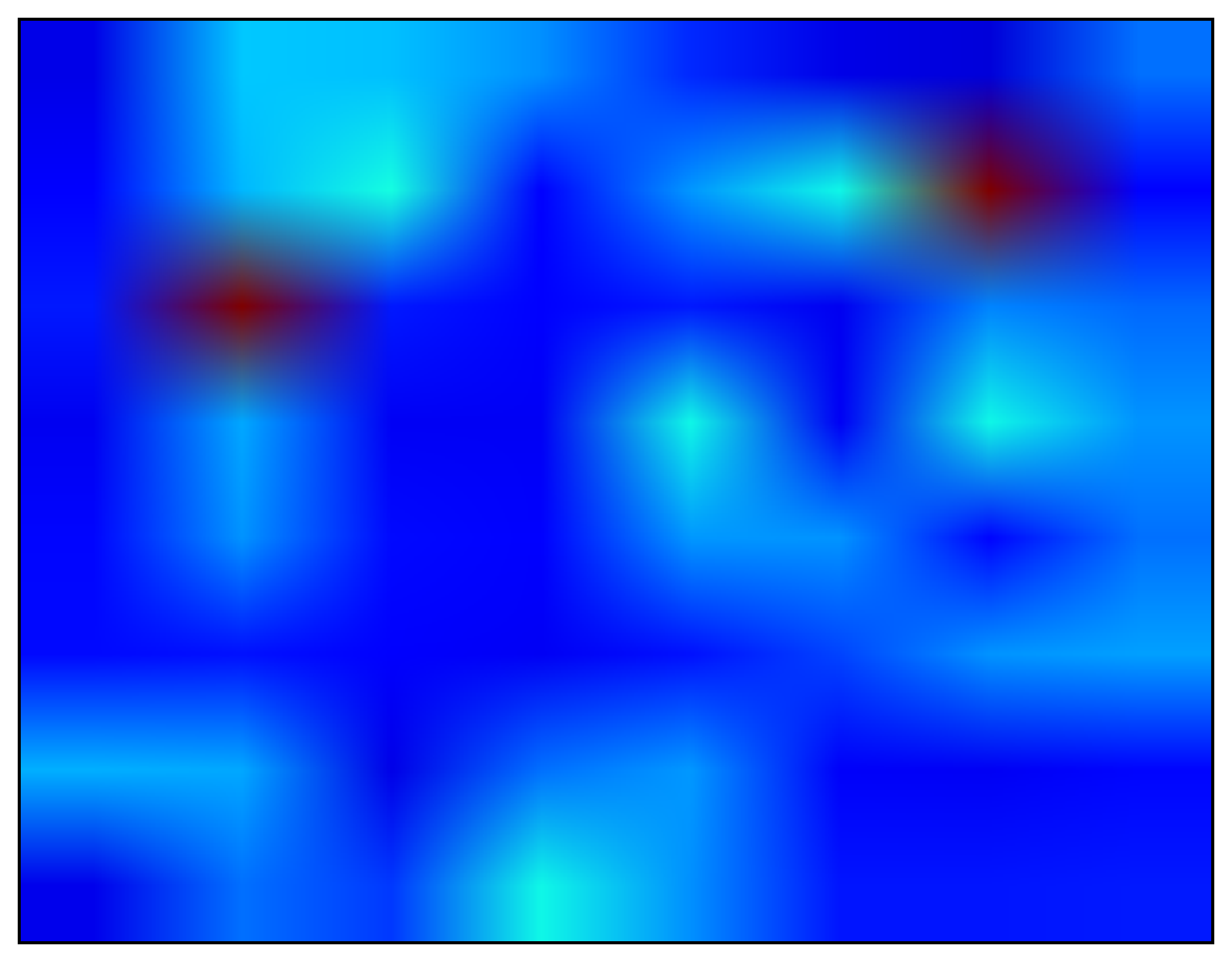}}
		
		\put(20, 53){\footnotesize{50}}
		\put(68, 53){\footnotesize{100}}
		\put(118, 53){\footnotesize{150}}
		\put(168, 53){\footnotesize{200}}
		\put(218, 53){\footnotesize{250}}			
		
		\put(0,2){\centering \includegraphics[width=47px, height=47px]{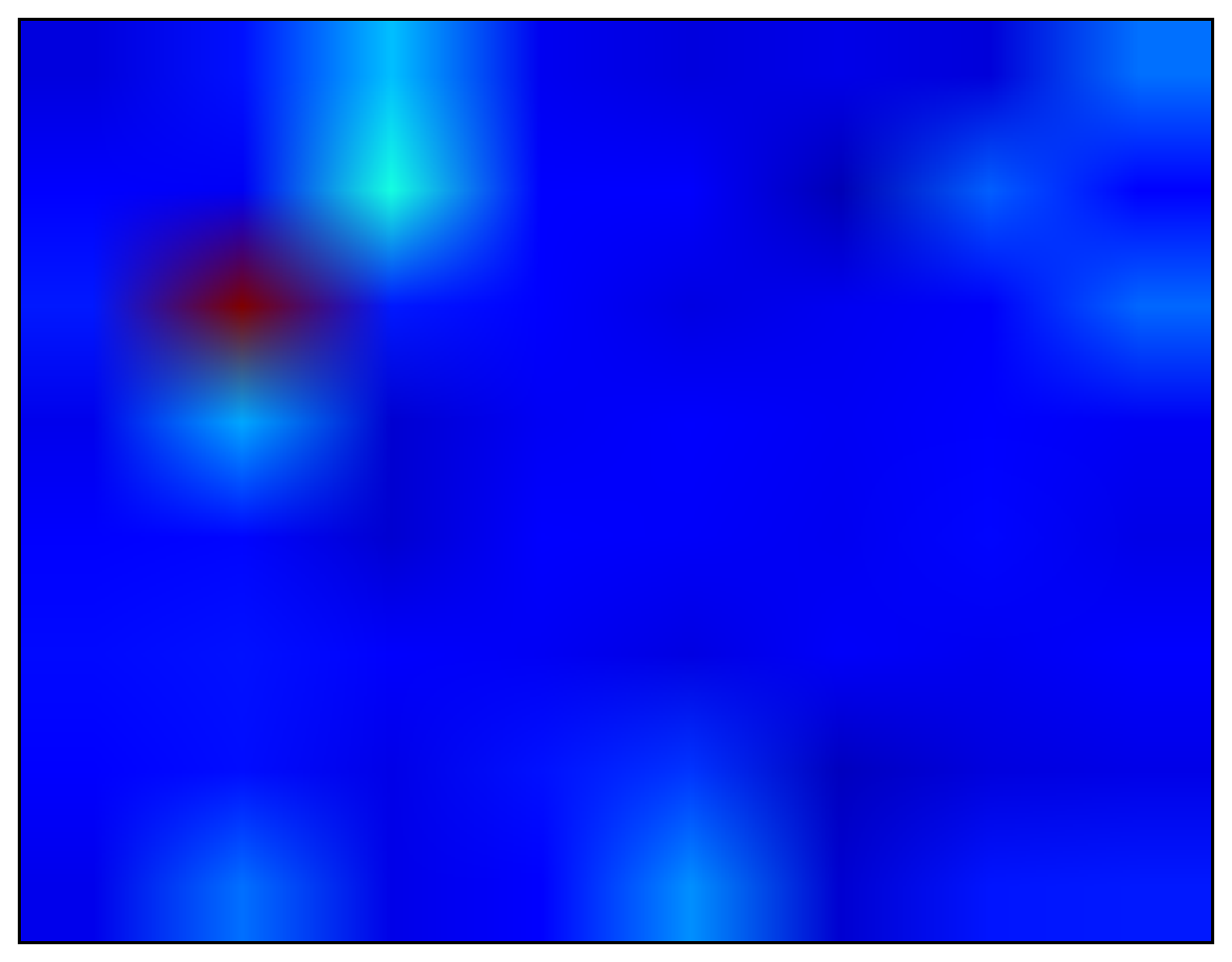}}
		\put(50,2) {\centering \includegraphics[width=47px, height=47px]{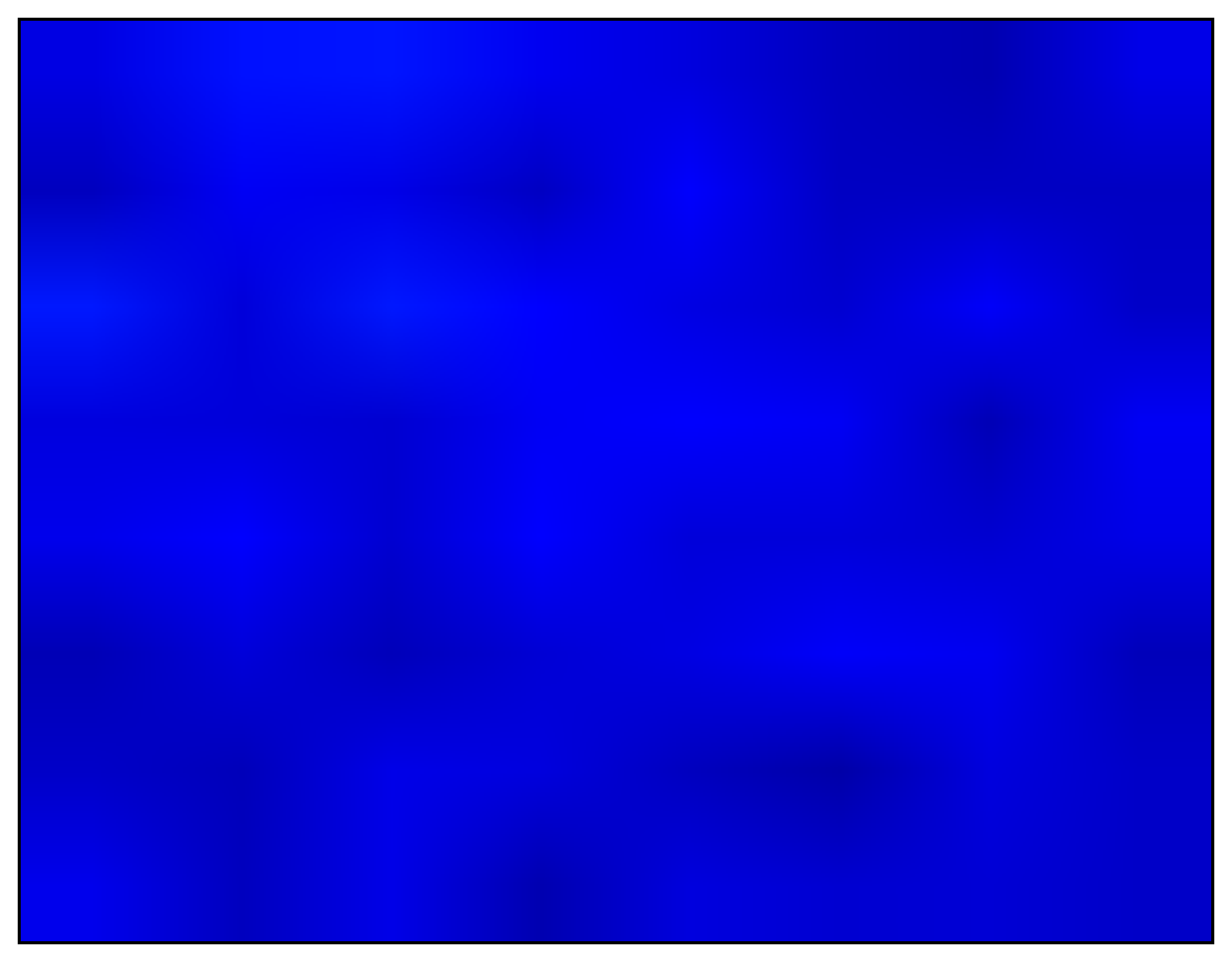}}
		\put(100,2){\centering \includegraphics[width=47px, height=47px]{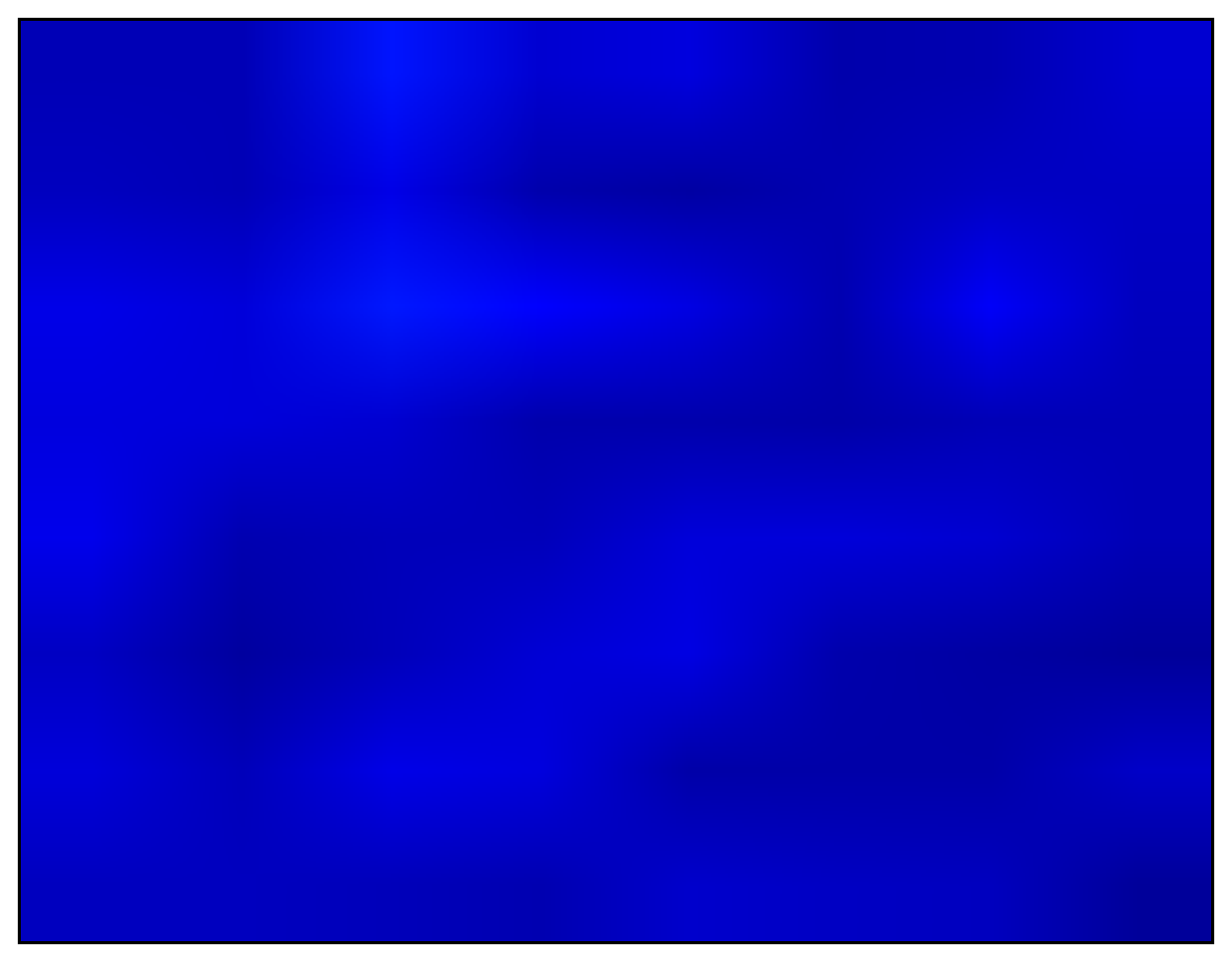}}
		\put(150,2){\centering \includegraphics[width=47px, height=47px]{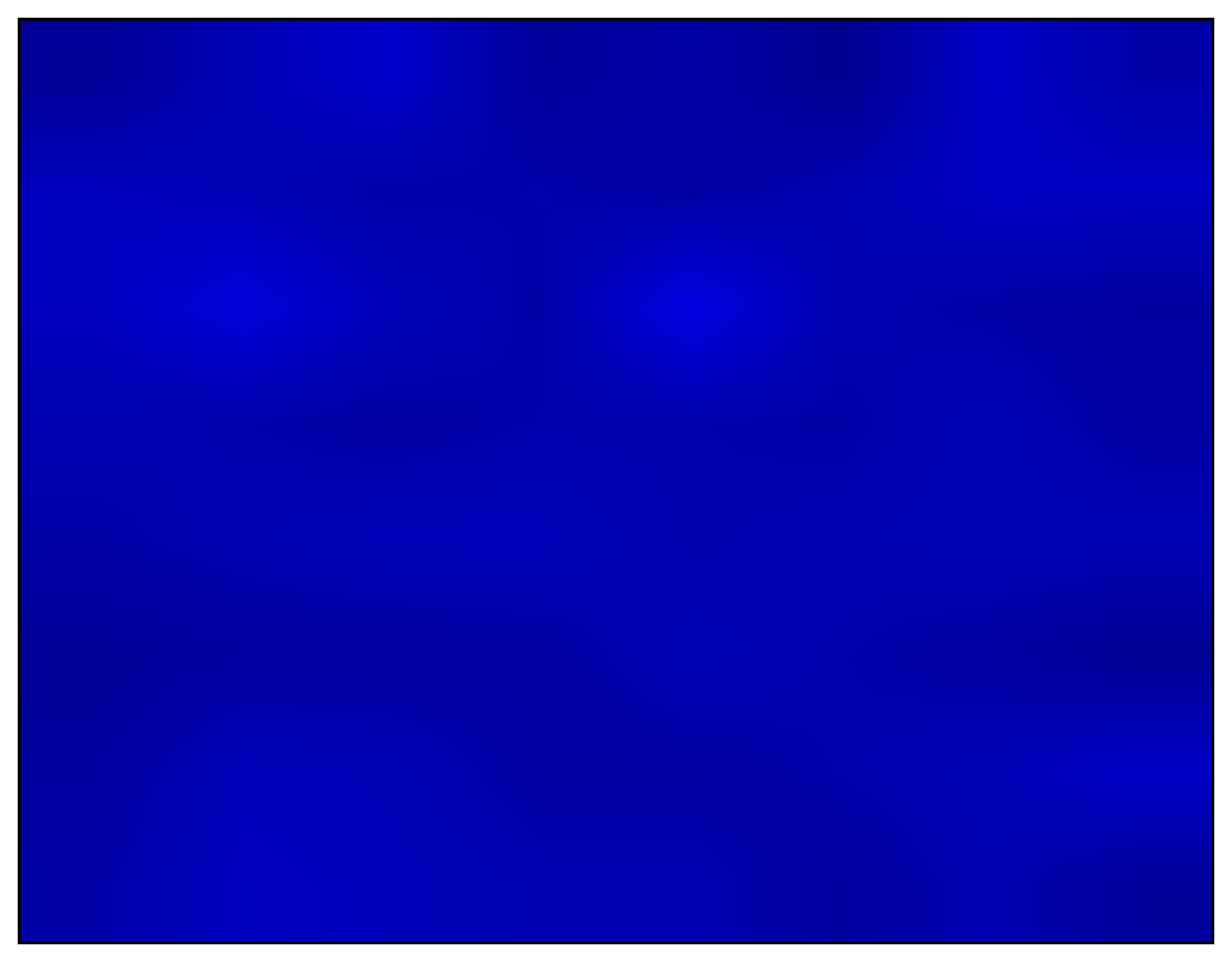}}
		\put(200,2){\centering \includegraphics[width=47px, height=47px]{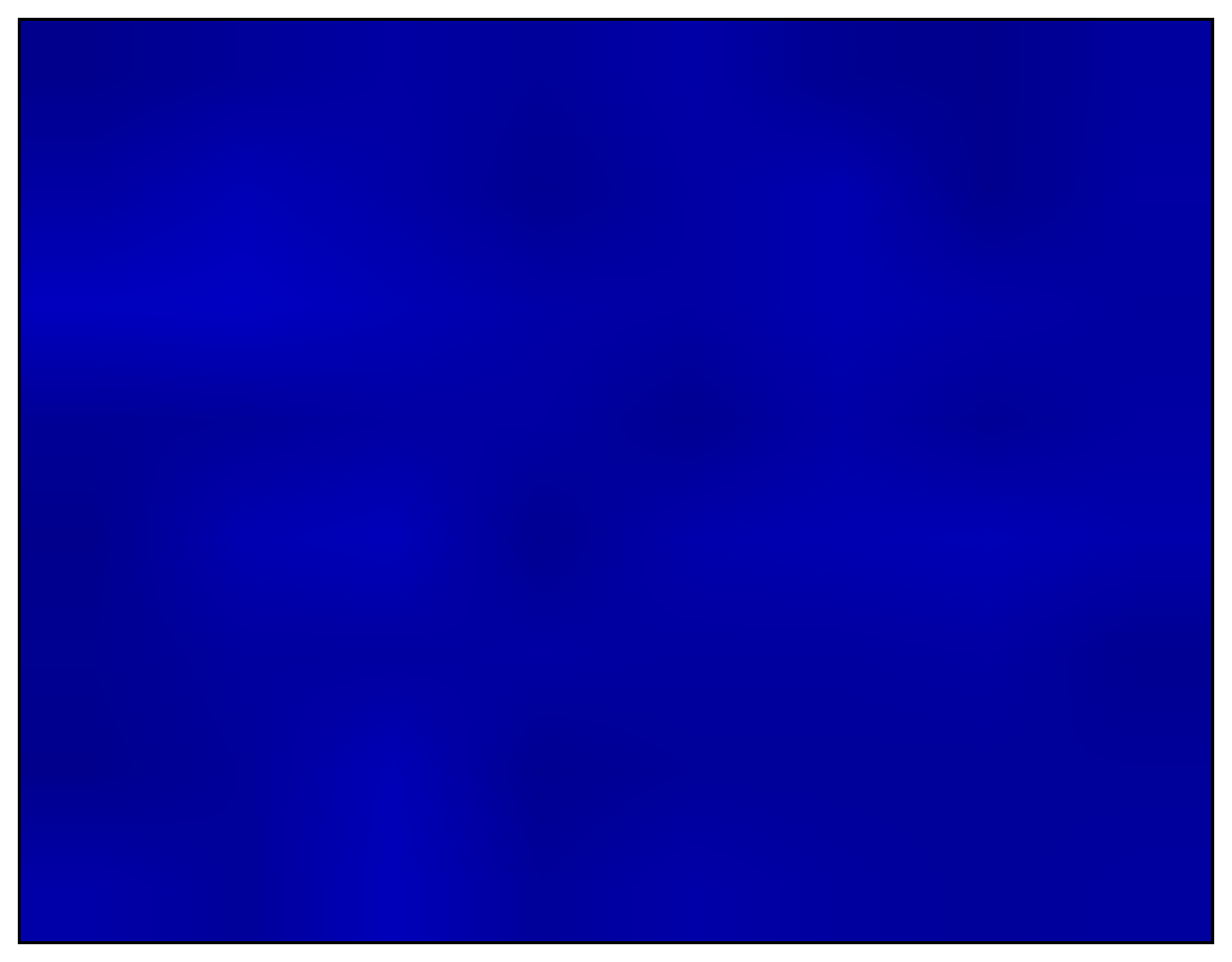}}
		
		\put(18, -5){\footnotesize{300}}
		\put(68, -5){\footnotesize{350}}
		\put(118, -5){\footnotesize{400}}
		\put(168, -5){\footnotesize{450}}
		\put(218, -5){\footnotesize{500}}
		\end{picture}
		\vspace{-5pt}\caption{Value functions of the same randomly selected run after a specific number of actions, indicating the current Q-values along state transition edges. Each row corresponds to a particular aspect node $s$ and each column corresponds to the action $a(\rho)$ that results in a particular $s'$. Only \textsc{Orbit} is considered in this image. The illustrated heatmap range is from (blue) $0.01$ to (red) $1.0$ (Best viewed in color). }
		\label{fig:value-functions}
		\vspace{-15pt}
	\end{figure}
	
\begin{figure}
		\centering
		\begin{picture}(250,113)
		\put(0,0){\includegraphics[width=0.49\textwidth]{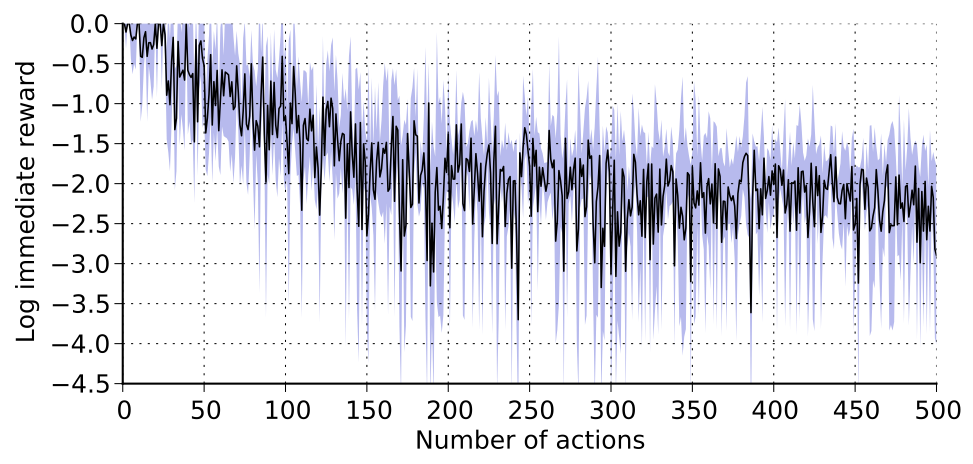}}
		\end{picture}
		\vspace{-15pt}
		\caption{Mean and one standard deviation over five trials for log immediate reward after the execution and update of each \textsc{Orbit} action taken in the first experiment for the proposed approach.}
		\label{fig:convergence}
		\vspace{-4.25pt}
	\end{figure}

	As hinted in Section~\ref{sect:intrinsic-motivation}, affordance structure learning does not converge to an optimal policy described by the resulting value function through value iteration. Instead, Figure~\ref{fig:value-functions} illustrates how the structure learning task \emph{consumes} value and depletes the intrinsic rewards as the number of executed actions increase. These value functions illustrated the Q-values along transition edges corresponding to states $s$ and actions $a(\rho)$. High values are attained initially when action parameters from aspect nodes have been inadequately unexplored. Immediate reward is consumed over time when actions are performed, condensing variance in transition distributions---such a phenomenon is illustrated in Figure~\ref{fig:convergence} in which the log immediate reward depletes as learning progresses. 
	
	\begin{figure}
		\centering
		\begin{picture}(250,122)
		
		\put(0,0){\includegraphics[width=0.49\textwidth]{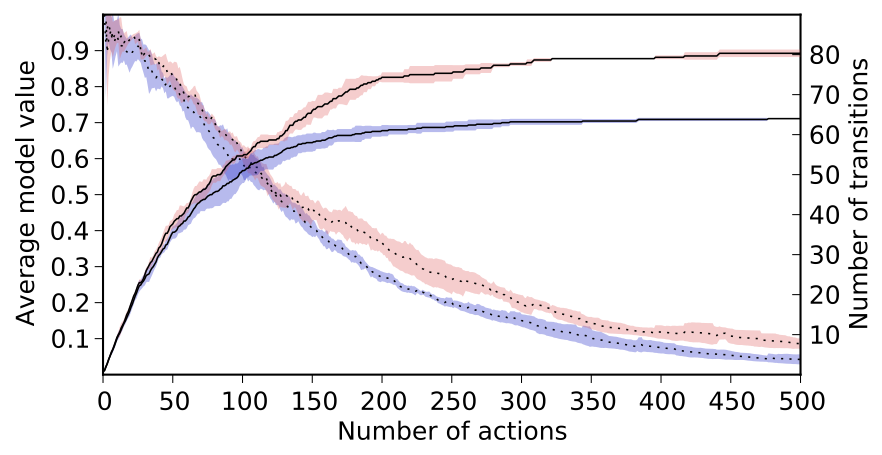}}
		
		\put(123,80.5){\includegraphics[width=0.125in,height=0.05in]{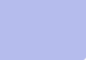}}
		\put(123,68){\includegraphics[width=0.125in,height=0.05in]{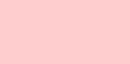}}
		
		\put(134,79.5){$\mathcal{A}=$\{\textsc{Orbit}\}}		
		\put(134,67){$\mathcal{A}=$\{\textsc{Orbit, Grasp}\}}	
		\end{picture}
		\vspace{-17pt}
		\caption{Increasing sensor modalities and action space. Mean and one standard deviation over five trials illustrating the average $Q$ value and the number of transitions discovered in the model using the proposed approach. The dotted lines correspond to average model value and solid lines describe the number of transitions in the affordance model (Best viewed in color).}
		\vspace{-15pt}
		\label{fig:multi-convergence}
	\end{figure}	
	
	The introduction of additional sensor modalities and actions results in slightly slower convergence, yet continues to discover all the transitions in the learned ATG. The result of over $100$ hours of simulation is illustrated in Figure~\ref{fig:multi-convergence}. As the the number of transitions discovered in the model increases, the likelihood of novelty diminishes---this is captured in the decreasing values in the model. The affordance model with an extended sensor modalities and actions $\mathcal{A}=\{\textsc{Orbit}, \textsc{Grasp}\}$ contains twelve aspect nodes and $80$ aspect transitions. Like in the first experiment, structure learning with the extended action set requires $300$--$400$ actions to produce a complete model. Other methods presented to learn affordances like OACs required a similar number of actions \cite{Ugur2014}. 
	
	\section{Discussion, Conclusion, and Future Work}
	
	This manuscript presents an intrinsically motivated structure learning approach to learn semi-permanent Markovian state representations of structures that are reusable in future (potentially partially-observable) tasks. The affordance representations learned here serves as a key component in belief-space object identification architectures \cite{Grupen2016}. These representations can be leveraged as forward models to predict how state distributions change in response to interaction. Despite success in the past using hand-crafted models of this type, the methods presented in this paper allows us to acquire them autonomously and encodes robot uncertainties and parameters that would otherwise be difficult to precisely hand define. Structure learning allows robots to build models themselves without supervision and promotes informed action selection, exploiting known structure and promoting a sense of discovery. Results demonstrate the acquisition of models that are significantly better than approaches that solely select random actions to learn from. With the proposed learning method, the transitions encode uncertainties in the form of distributions derived from the properties of the embodied system and its interaction with the affordances of the world---as such, methods like this allow for a possibility for not only even finer-grained error detection, but also, support error \emph{correction} in many cases, producing a more general and robust representation for planners and belief-space architectures. Future work looks into extending the action space and modalities further and investigating methods to take a learned affordance representation and decompose known aspect nodes to incorporate new sensory information while preserving learned transition dynamics. We believe that autonomously learning affordance representations as forward models with more complex actions and modalities allows for a richer set of future solvable tasks. Furthermore, despite the growing complexity of the affordance representation as more actions are introduced, we believe that enriched models reduce the complexity in model-referenced planning, thus reducing planning time and the number of rollouts necessary to solve future tasks. 
	
	

	
	\section*{Acknowledgements}
	The authors thank Michael W. Lanighan for his initial contributions. We also thank Mitchell Hebert and Samer Nashed for their feedback on this manuscript. This material is based upon work supported under Grant NASA-GCT-NNX12AR16A. Any opinions, findings, conclusions, or recommendations expressed in this material are solely those of the authors and do not necessarily reflect the views of the National Aeronautics and Space Administration.
	
	\bibliographystyle{IEEEtran}
	\footnotesize{\bibliography{developmental}

\end{document}